\documentclass[journal]{vgtc}                     % final (journal style)
\def \etal {{\emph{et al}.\thinspace}}
\def \eg {{\emph{e.g}.\thinspace}}

\def \ie {{\emph{i.e}.\thinspace}}
\newcommand{\sys}{{DeepVIS}\xspace}
\newcommand{\data}{{nvBench-CoT}\xspace}
\newcommand{\model}{{NL2VIS-CoT}\xspace}
\newcommand{\task}{{NL2VIS}\xspace}
\newcommand{\revise}[1]{{\color{black}{#1}}}

\newcommand{\E}[1]{E{#1}}

\newcommand{\sql}[1]{%
  {\ttfamily #1}%
}
\onlineid{0}

\vgtccategory{Research}

\vgtcpapertype{algorithm/technique}

\title{\sys: Bridging Natural Language and Data Visualization Through Step-wise Reasoning}

\author{%
   \authororcid{Zhihao Shuai}{0009-0004-0110-3113},
   \authororcid{Boyan Li}{0009-0009-8391-4687},
   \authororcid{Siyu Yan}{0009-0003-3419-4608},
   \authororcid{Yuyu Luo}{0000-0001-9530-3327}, and
   \authororcid{Weikai Yang}{0000-0002-6520-1642}
 }
\authorfooter{
   \item
   	Zhihao Shuai, Boyan Li, Siyu Yan, Yuyu Luo, Weikai Yang are with the Hong Kong University of Science and Technology (Guangzhou), Guangzhou, China. Yuyu Luo and Weikai Yang are also with the Hong Kong University of Science and Technology, Hong Kong SAR, China. Zhihao Shuai and Boyan Li are joint first authors. Weikai Yang is the corresponding author. E-mail: \{zhihaoshuai@, bli303@connect., syan195@connect., yuyuluo@, weikaiyang@\}hkust-gz.edu.cn
 }

\abstract{%
Although data visualization is powerful for revealing patterns and communicating insights, creating effective visualizations requires familiarity with authoring tools and often disrupts the analysis flow.
While large language models show promise for automatically converting analysis intent into visualizations, existing methods function as black boxes without transparent reasoning processes, which prevents users from understanding design rationales and refining suboptimal outputs. 
To bridge this gap, we propose integrating Chain-of-Thought (CoT) reasoning into the Natural Language to Visualization (\task) pipeline.
First, we design a comprehensive CoT reasoning process for \task and develop an automatic pipeline to equip existing datasets with structured reasoning steps.
Second, we introduce \data, a specialized dataset capturing detailed step-by-step reasoning from ambiguous natural language descriptions to finalized visualizations, which enables state-of-the-art performance when used for model fine-tuning.
Third, we develop \sys, an interactive visual interface that tightly integrates with the CoT reasoning process, allowing users to inspect reasoning steps, identify errors, and make targeted adjustments to improve visualization outcomes.
Quantitative benchmark evaluations, two use cases, and a user study collectively demonstrate that our CoT framework effectively enhances \task quality while providing insightful reasoning steps to users.
}
\teaser{
  \centering
  \includegraphics[width=\linewidth, alt={A view of a city with buildings peeking out of the clouds.}]{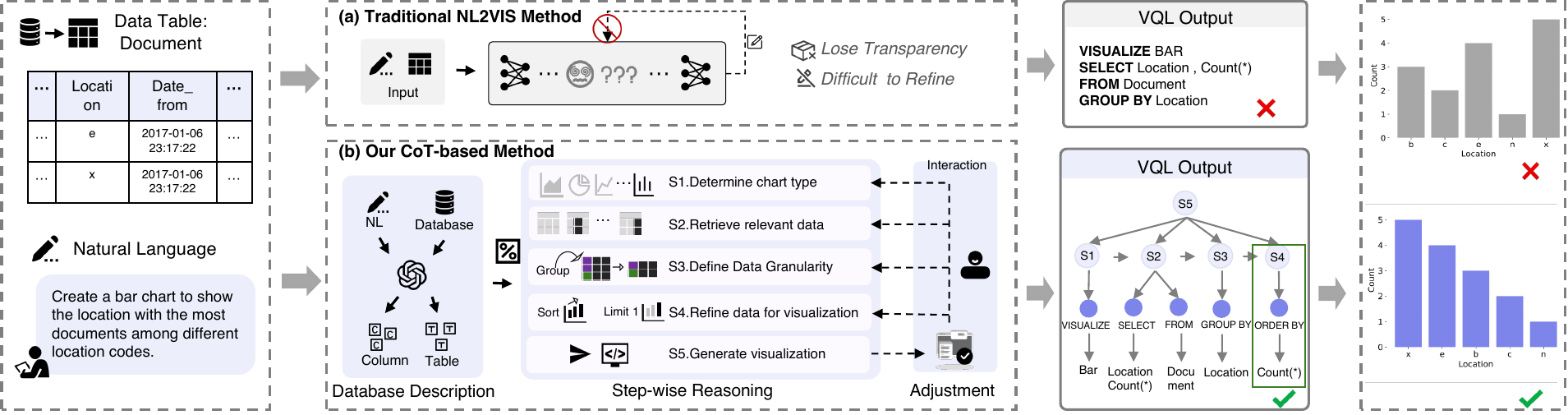}
  \caption{%
Comparing traditional \task methods with our method:
(a) Traditional methods often function as black boxes, making it difficult for users to interpret design rationales or refine suboptimal results.
(b) Our method leverages CoT reasoning, which enhances both model performance and transparency, enabling users to better understand the model decision and interactively refine results.
  }
  \label{fig:teaser}
}

\graphicspath{{figs/}{figures/}{pictures/}{images/}{./}} % where to search for the images
\usepackage{tabu}                      % only used for the table example
\usepackage{booktabs}                  % only used for the table example
\usepackage{lipsum}                    % used to generate placeholder text
\usepackage{mwe}                       % used to generate placeholder figures
\usepackage{mathptmx}                  % use matching math font

\begin{document}

%%%%%%%%%%%%%%%%%%%%%%%%%%%%%%%%%%%%%%%%%%%%%%%%%%%%%%%%%%%%%%%%
%%%%%%%%%%%%%%%%%%%%%% START OF THE PAPER %%%%%%%%%%%%%%%%%%%%%%
%%%%%%%%%%%%%%%%%%%%%%%%%%%%%%%%%%%%%%%%%%%%%%%%%%%%%%%%%%%%%%%%

% \firstsection{Introduction}

\maketitle

\section{Introduction}

Data visualization serves as a powerful tool in the data analysis pipeline, enabling effective exploration, pattern recognition, and the communication of insights~\cite{nl2sql_survey,deepeye_tkde,DBLP:journals/pvldb/LiLCLT24,linenet,statqa}.
Despite its critical importance, creating high-quality visualizations remains a challenging task that typically requires both specialized knowledge of visualization principles and proficiency with complex authoring tools~\cite{luo2021ncnet,tian2024chartgpt,podo2024agnostic}.
This technical expertise requirement presents a barrier for many data analysts, particularly those without formal training in visualization design~\cite{wu2024automated}. 
Even for experienced analysts, the process of switching between analysis environments and authoring tools forces them to mentally translate their analysis intent into the specific commands or parameters required by authoring tools, which disrupts analysis flow and \revise{negatively affects their productivity}~\cite{wu2022nl2viz}.

Given these challenges, there is a compelling need for systems that can automatically transform natural language descriptions of analysis intent into appropriate visualizations.
Such systems would enable analysts to remain focused on their primary task of data exploration and insight discovery, rather than becoming distracted by the mechanism of visualization creation.
Recent advancements in large language models (LLMs) have demonstrated considerable promise for the Natural Language to Visualization (\task) task, exhibiting remarkable capabilities to interpret user intent and generate corresponding visualization specifications~\cite{wu2024automated,li2024prompt4vis}.
\revise{Such systems would enable analysts to remain focused on their primary task of data exploration and
insight discovery, rather than becoming distracted by the mechanics of visualization creation~\cite{podo2024agnostic}.}
However, despite their potential, current methods function largely as black boxes, processing natural language inputs and producing visualization outputs without exposing the intermediate reasoning steps \revise{(Fig.~\ref{fig:teaser}(a))}.
This opacity creates several critical limitations:
First, users cannot understand how or why specific visualization choices were made, which reduces their trust in the model outputs.
Second, when faced with suboptimal outputs, users lack visibility into specific points of failure within the reasoning process, which hinders their ability to pinpoint and rectify problems. 
Third, users miss the valuable chance to learn from the decision-making process of the model, which could enhance their own visualization expertise. 

To address these limitations, we propose integrating Chain-of-Thought (CoT) reasoning into the \task pipeline \revise{(Fig.~\ref{fig:teaser}(b)),} \revise{which encourages LLMs to break down complex problems into intermediate steps~\cite{wei2022chain, chen2025towards}}.
First, we design a comprehensive CoT reasoning process for the \task task and develop an automatic pipeline to equip existing \task datasets with structured CoT reasoning steps.
Based on this, we introduce a new dataset, \data, which captures the detailed step-by-step reasoning processes that connect ambiguous natural language descriptions to finalized visualizations, mimicking the design process of experienced analysts.

We then train an \task model on this dataset that explicitly incorporates CoT reasoning.
Unlike existing black-box methods, our model exposes the intermediate reasoning steps that guide the transformation from natural language input to visualization output.
This transparency not only enables the model to achieve state-of-the-art performance but also enables users to understand the rationale underlying visualization choices and identify potential areas for improvement. 
We also develop \sys, an interactive visual interface that tightly integrates with the CoT reasoning process.
This interface empowers users to inspect each step of the reasoning chain, identify potential errors or suboptimal decisions, and strategically intervene by modifying specific reasoning components to enhance the final visualization.
By facilitating this form of human-AI collaboration, our system effectively combines the efficiency of automated visualization generation with the judgment of human analysts.

We evaluate our method through quantitative benchmark evaluation, two use cases, and a user study.
The results demonstrate that our CoT framework significantly improves the quality and accuracy of \task transformations compared to black-box methods. Furthermore, user feedback indicates that the transparent reasoning process substantially enhances trust in the model outputs while providing valuable learning opportunities that enable users to refine their own visualization expertise.
To summarize, our contributions are threefold:

\begin{itemize}[noitemsep,topsep=0pt]
  \item We design a comprehensive CoT reasoning process for the \task task, which is used to guide the process of enhancing \task datasets with structured reasoning steps. 
  \item We introduce and curate \data, an \task dataset with detailed reasoning steps, and achieve state-of-the-art performance by fine-tuning models on this dataset.
  \item We develop a visual interface that tightly integrates with the CoT reasoning process, which allows users to understand the underlying reasoning and make targeted adjustments.
\end{itemize}

\section{Related Work}

\subsection{\revise{Chain-of-Thought (CoT)}}
\revise{While LLMs have demonstrated impressive performance, they often face challenges when processing complex problems.
The CoT technique addresses this by guiding LLMs to decompose complex problems into a series of intermediate steps, thereby improving reasoning and transparency~\cite{wei2022chain,feng2023towards}.
The effectiveness of CoT is significantly influenced by the design of these intermediate steps.
Therefore, several strategies have been developed to generate high-quality reasoning steps, such as zero-shot CoT that instructs the model to generate reasoning steps~\cite{kojima2022large,alpha-sql}, few-shot CoT that provides the model with a few examples of similar problems with reasoning steps~\cite{wei2022chain,aot}, Tree-of-Thought that enables exploration of multiple reasoning paths and dynamic adjustment through backtracking~\cite{yao2023tree}, and interactive construction and involve human in crafting the chain~\cite{wu2022promptchainer,wu2022ai}, 
For a comprehensive overview, we refer readers to recent surveys~\cite{chu2023navigate,chen2025towards}.
In this work, we systematically analyze \task and design a tailored CoT reasoning pipeline, which effectively guides LLMs through the complex stages of visualization generation and improves performance.}

\subsection{\task}
Based on their underlying principles and implementation mechanisms, existing \task methods can be classified into three categories: Rule-based methods, translation-based methods, and LLM-based methods.

The early efforts in \task are rule-based methods, which utilize predefined rules to analyze natural language queries and transform them into a set of predefined visualization templates~\cite{ sun2010articulate,narechania2020nl4dv,yu2019flowsense,luo2018deepeye, gao2015datatone,setlur2016eviza,srinivasan2023bolt,chen2022pi2}.
Articulate~\cite{sun2010articulate} is one of the pioneering works, which utilizes a parser to tag each word and then classify the whole query into different analysis tasks.
A suitable chart is then generated based on the analysis tasks and the data to be visualized.
Later efforts focus on improving performance in handling ambiguity in natural language input. 
For example, NL4DV~\cite{narechania2020nl4dv} simultaneously considers syntactic and semantic similarity to identify data attributes referenced in the query.
While these methods provided an initial framework for \task, they were often constrained in their ability to handle complex or unforeseen queries and required substantial manual effort in defining the rules and templates.

With the advancement of machine learning translation, translation-based methods are proposed to solve \task by treating it as a translation problem between human languages and visualization languages~\cite{luo2021ncnet,luo2021synthesizing,nvbench2}.
Specifically, they often utilize sequence-to-sequence models (\eg, recurrent neural network or transformer) to encode the natural language query into a hidden representation and then decode it into a visualization specification.
A representative work in this category is ncNet~\cite{luo2021ncnet}, which is a Transformer-based model incorporating visualization-aware optimizations to enhance the translation process and the quality of the generated visualizations.
To better \revise{address the insufficient accuracy of from-scratch generation methods,} RGVisNet~\cite{song2022rgvisnet} leverages a hybrid retrieval-generation method to enhance results by retrieving the most relevant queries from existing data.
While these methods offer greater flexibility and the ability to learn visualization-specific knowledge, they sometimes still struggle in understanding complex and ambiguous natural language queries.

The recent surge in the capabilities of LLMs has led to their widespread adoption in \task.
\revise{To comprehensively assess the capabilities and limitations of LLMs, researchers have conducted extensive evaluation~\cite{vazquez2024llms,hong2025llms,10.2312:eurova.20241118,pandey2025benchmarking,chen2024viseval,kim2023good}.
For example, Vazquez~\etal\cite{vazquez2024llms} conducted systematic experiments analyzing LLM performance in NL2VIS across different aspects, including chart generation, library adaptation, and visual variable configuration.
Chen~\etal\cite{chen2024viseval} constructed VisEval, a comprehensive NL2VIS benchmark that evaluates multiple LLMs to identify common challenges and provide insights for future research directions.
Beyond evaluation,} researchers have also explored methods to better harness the strong language understanding and generation abilities of LLMs through techniques like in-context learning~\cite{li2024prompt4vis,maddigan2023chat2vis,chen2022nl2interface,kavaz2023chatbot} and supervised fine-tuning~\cite{xie2024haichart, tian2024chartgpt,podo2024v}.
For example, \revise{LLM4VIS~\cite{kahng2024llm} employs a few-shot approach to guide the model in recommending appropriate visualization types for the test data,} while Prompt4VIS~\cite{li2024prompt4vis} retrieves similar questions and the groundtruth answers as input \revise{to enhance model performance.}
\revise{To further help models make correct reasoning, recent work like} ChartGPT~\cite{tian2024chartgpt} \revise{and  V-RECS~\cite{podo2024v}} decompose the visualization generation process into a series of sub-tasks that the LLM addresses sequentially.
In contrast to existing approaches, our method provides a detailed reasoning process for each step, which not only effectively boosts model performance but also enhances transparency.\looseness=-1

\subsection{Visualization for Human-LLM Collaboration}
Visualization has been proven to be an effective way to help users harness the power of LLMs and achieve different tasks~\cite{yuan2021survey,wang2021survey,yang2023survey,yang2024foundation,Liu2025,wang2024visual}.
Based on the purpose of the visualization, existing work can be classified into two categories: visualizations for enhancing LLM inputs and visualizations for enhancing LLM outputs.

A significant body of work has explored how visualization aids users in crafting effective prompts and achieving better performance. 
For example, Strobelt~\etal\cite{strobelt2022interactive} proposed PromptIDE, a tool designed to assist users in constructing prompts for text classification tasks.
It allows users to construct multiple prompt variations, compare their performance, and iteratively refine them based on quantitative feedback.
For complicated tasks that require complex prompts to unlock the potential of LLMs, PromptChainer~\cite{wu2022promptchainer} and AI Chains~\cite{wu2022ai} help users decompose complicated into smaller, more manageable sub-tasks, thereby simplifying the creation of prompts.
In the domain of text-to-image generation, PromptMagician~\cite{feng2023promptmagician} supports users to efficiently explore and compare the prompts and corresponding generated images retrieved from a database.
This visual exploration offers valuable guidance and hints for refining user prompts.
PromptCharm~\cite{wang2024promptcharm} focuses on iteratively improving generated images through multimodal prompting by allowing users to adjust the attention given to specific keywords within the prompt.
PromptMap~\cite{promptmap} introduces an intuitive spatial interface for image generation, allowing users to manipulate prompts via interactive 2D layouts (\eg, grids or graphs) instead of text.
For multimodal reasoning tasks, POEM~\cite{he2024poem} visualizes interaction patterns across modalities at different granularities.
This multi-level model understanding empowers users to refine prompts in a more interpretable and controllable manner.\looseness=-1

Another prominent area of research focuses on facilitating human-LLM collaboration by enabling users to understand and steer the output of LLMs.
For example, InsightLens~\cite{weng2024insightlens} provides a structured and accessible way for users to record, organize, and revisit these insights, enhancing the overall efficiency and value of LLM-powered data analysis workflows.
Patchview~\cite{chung2024patchview} uses a tangible visual metaphor that enables writers to intuitively guide the LLM in generating story world elements, fostering a more natural and expressive co-creation experience.
TaleBrush~\cite{talebrush} employs
line-sketching interactions along with a GPT-based language model to support writers in dictating character fortune plots in line with the creative goals of the writers.
The most relevant one is
WaitGPT~\cite{xie2024waitgpt}, which facilitates programmers to understand and verify the code generated by LLMs for data analysis tasks.
It transforms the LLM-generated code into an interactive node-link diagram that updates in real time, visually representing data operations and their intermediate states.
Users can efficiently monitor the analysis process, understand the logic behind the generated code, and even modify specific operations directly within the visual interface.
Our work aligns with the second category, focusing on making the reasoning steps of LLMs more transparent and controllable. 
However, unlike WaitGPT, which transforms generated code into code in a post-analysis manner, we first design a comprehensive CoT reasoning process tailored for \task, providing a step-by-step explanation of how natural language descriptions are translated into finalized visualizations.
This structured process not only boosts performance in \task but also provides better transparency.\looseness=-1

\section{Design of the CoT Process for \task}
The key to incorporating the CoT process into the \task pipeline lies in understanding how analysts design appropriate visualizations to fulfill their analysis intent.
To develop a CoT process that effectively captures visualization design reasoning, we conducted a comprehensive literature review and expert interviews, resulting in a structured five-stage CoT process that is suitable for \task tasks.

\begin{figure*}[htbp]
  \centering
  \includegraphics[width=\textwidth]{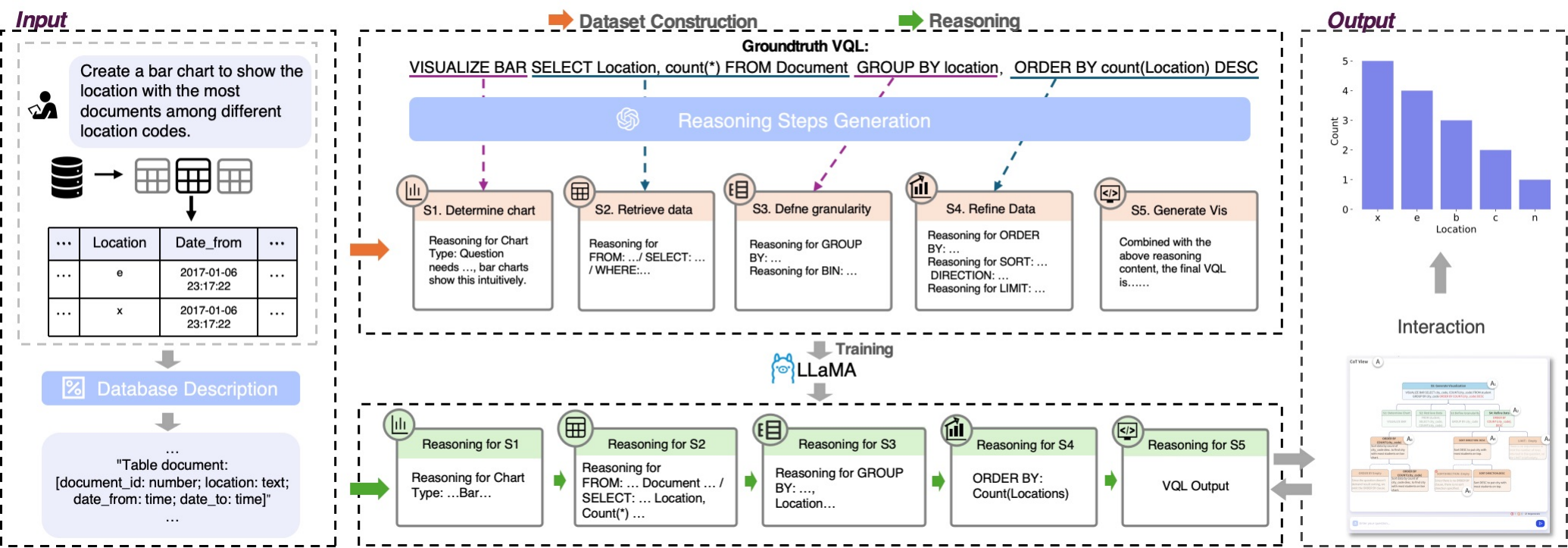}
  \vspace{-6mm}
    \caption{The pipeline of DeepVIS Framework}
    \label{fig:pipeline}
\end{figure*}

\subsection{Literature Review and Expert Interviews}
We began by conducting a comprehensive review of existing literature, focusing on frameworks that describe the systematic progression from analysis intent to visualizations~\cite{munzner2009nested,wang2022towards,wilkinson2011grammar} and \task~\cite{tian2024chartgpt,shen2022towards}.
The nested model proposed by Munzner~\cite{munzner2009nested} was particularly influential, which emphasizes progression from domain problem characterization to data abstraction and visual encoding.
In addition, prior research \cite{tian2024chartgpt} has demonstrated decoupling the \task task from aspects such as data transformation and visualization notably boosts visualization accuracy. 
To complement our theoretical analysis, we conducted semi-structured interviews with four visualization experts (\E1-\E4) to understand their real-world visualization authoring workflows.
\E1 is a senior researcher with 8-year experience in data visualization;
\E2 is a data analyst familiar with multiple visualization tools;
\E3 and \E4 are Ph.D. students with 4 and 3 years visualization experience, respectively.
None of them are co-authors of this work.
Using a think-aloud protocol, experts verbalized their thought processes while designing appropriate visualizations.
Each interview lasted between 40 and 55 minutes.

\subsection{Initial Three-Stage Process}
Based on our literature review and expert interviews, we initially identified three key stages in the visualization authoring process:

\noindent\textbf{Understand analysis intent and determine chart type}.
In this initial stage, analysts first understand the visualization goal and select an appropriate chart type.
As E1 stated, ``I will first quickly examine a few exemplar data and understand the visualization goal, then I can choose the most suitable visualization type to fulfill my analysis need.''
E4 further elaborated, ``Even if the exact data processing is not immediately clear, I usually start by determining the chart type and identifying the primary axes.
It will provide a clear picture of what the visualization might reveal.''
These insights align with findings from Wang~\etal\cite{wang2022towards}, which noted that analysts preferred a top-down method, starting with determining the chart type and then specifying other configuration details.\looseness=-1

\noindent\textbf{Prepare data for visualization}.
After determining the chart type, all experts agreed they would start processing raw data to suit the specific requirements of that chart.
As E3 explained, ``Once I know I am creating a bar chart to compare categories, I need to make sure my data is aggregated correctly, since there is no point showing individual data points and only the summary statistics matter.''
This data processing stage usually involves filtering the data to focus on relevant subsets, aggregating data to summarize key trends or patterns, and transforming the data by calculating new fields.
This acts as a crucial bridge between the raw data and the visual representation, ensuring that the data is structured and formatted in a way that is suitable for the selected chart type.\looseness=-1

\noindent\textbf{Convert prepared data into visual elements}.
Following data preparation, the final stage involves mapping the data into visual channels appropriate to the selected chart type, such as the height of bars or the area of shapes.
E2 described this process as ``assigning the right variables to the right visual properties, such as putting the independent variable on the x-axis and the dependent on the y-axis.''
E4 also noted that he would usually make some adjustments after examining the results, such as removing bars with very low values to reduce the number of bars.
In short, this stage aims to effectively and accurately visualize the prepared data and clearly communicate the intended message to the audience.
\looseness=-1

\subsection{Refined to Five-Stage CoT Process for \task}
After summarizing these three stages, we presented them to our experts for validation and collected their feedback.
While they generally agreed with this pipeline, they pointed out that they could be further refined for the \task task, especially in data preparation.
E3 stated, ``When analyzing trends over time, I usually experiment with different temporal granularities, like daily, monthly, or yearly.
This does not change the underlying data I visualize but only changes the granularity of analysis.''
E2 elaborated, ``This is essentially the binning operation, which is commonly used in exploratory data analysis and affects how patterns emerge from the data.''
Furthermore, E4 suggested splitting the data preparation into three steps: identifying relevant data, determining proper analysis granularity, and implementing necessary modifications to enhance visualization effectiveness.
He provided an example:
``When visualizing average income across different countries, we first extract data on income and nationality, then calculate the average income per country, and finally select and sort the countries and display the top 10 or 20 countries for visualization.
Once these three steps are completed, the data is properly prepared for effective visualization.''

Based on these feedback, we refined the CoT process structure to a five-stage process:
\begin{enumerate}[noitemsep,topsep=0pt]
    \item[S1] Determine chart type: Select the most appropriate visualization method based on the data and analysis intent.
    \item[S2] Retrieve relevant data: Identify and extract the specific data attributes required for the visualization.
    \item[S3] Define data granularity: Establish the appropriate granularity for the data visualization.
    \item[S4] Refine data for visualization: Apply transformations such as filtering and sorting to prepare the final data for optimal visualization. 
    \item[S5] Generate visualization: Configure and generate the final visualization to effectively communicate the intended insights.
\end{enumerate}

\subsection{Validation of the Five-Stage Process}
To validate our refined five-stage process, we conducted a follow-up evaluation with the same group of experts.
We presented them with five diverse analysis tasks and asked them to describe their visualization design process using our framework.
The experts successfully applied the five-stage process to all scenarios, confirming its completeness and flexibility.
E4 noted, ``This structure is comprehensive yet simple enough to apply across different types of visualization tasks.
It captures the key decision points without being overly prescriptive.''
E2 added, ``What I particularly like about this framework is that it explicitly details decisions often implicitly made by experienced visualizers, which could benefit novice training and automation.''
The validation confirmed that our five-stage CoT process effectively captures the essential reasoning steps in visualization design while remaining adaptable to diverse analytical scenarios.

\section{Construction of \data}
Even with the identified CoT process, we still need sufficient training data to guide models in learning the reasoning steps for \task.
However, it would be prohibitively expensive to manually create the whole reasoning process, which requires an immense amount of time and expert effort.
% because detailed, step-by-step reasoning chains require an immense amount of time and expert effort.
To tackle this issue, we develop an automatic pipeline that augments existing \task datasets with structured CoT reasoning steps.
\revise{Unlike the few-shot prompting approach~\cite{wu2024automated}, our pipeline significantly enhances the model's capability to reason about chart-specific patterns and SQL-syntax patterns.}
Fig.~\ref{fig:pipeline} shows our detailed pipeline, which consists of two modules: the database description module and the reasoning steps generation module.

For clarity, we demonstrate our pipeline using nvBench~\cite{luo2021synthesizing}, a widely used \task dataset, but our pipeline can be easily adapted to other \task datasets.
In nvBench, each training sample contains a table (\eg, \sql{Faculty}), a natural language query (\eg, ``Compute the total number of rank across rank as a pie chart''), and the corresponding Visualization expressed in Visualization Query Language (VQL) (\eg, \sql{VISUALIZE Pie SELECT Rank, COUNT(Rank) FROM Faculty GROUP BY Rank}).
This VQL can be executed, transformed to Vega-lite specifications, and rendered to obtain the visualizations.

\subsection{Database Description Module}
The database description module aims to enhance input to provide essential details for subsequent reasoning steps.
Our expert interviews revealed that analysts typically scan multiple table rows at the beginning to gain a foundational understanding of data.
Our experiments also confirmed that adding comprehensive database descriptions significantly improves reasoning accuracy and reduces errors (Sec.~\ref{subsec:ablation}).
To mirror this common practice of examining several table rows, we implemented a template-based augmentation method with two key subcomponents:

\noindent\textbf{Schema description}.
Given a table, we generate a comprehensible schema that captures essential elements, including table names, column names, and their data types.
For instance, when processing the \sql{faculty} table, our method formats columns using the conventional \sql{column name:value type} pattern, producing entries such as \sql{facid:number}, \sql{fname:text}, and \sql{rank:text}.

\noindent\textbf{Value sampling}.
In addition to the \revise{schema description},
it is also necessary to examine sample values for accurate reasoning.
For instance, a \sql{rank:text} column might contain either full names like \sql{Associate Professor} or abbreviation like \sql{AssocProf.}
Without concrete examples, it is impossible to determine whether \sql{WHERE rank=Associate Professor} or \sql{WHERE rank=AssocProf} is the correct filtering condition.
Therefore, we incorporate representative value samples in the input.
To ensure inclusion of relevant samples, we use GPT-4o-mini to process both the natural language query and database schema to identify relevant columns and incorporate appropriate samples accordingly.

\begin{figure*}[!t]
\includegraphics[width=\textwidth]{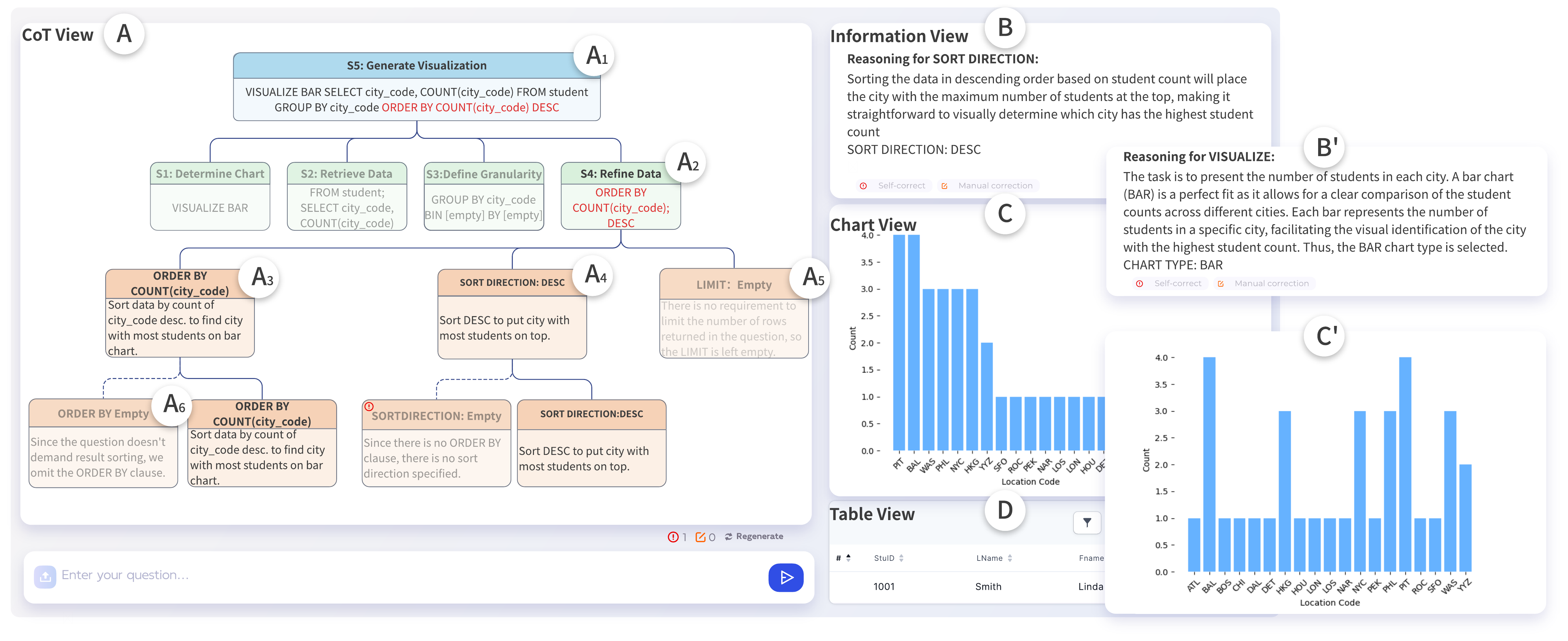}
\vspace{-6mm}
\caption{The interface of \sys. The CoT view (A) provides a structured overview of the reasoning process and allows users to interactively refine it, while three supporting views (B-D) complement the CoT view during the analysis.}
\label{fig:interface}
\end{figure*}

\subsection{Reasoning Steps Generation Module}
After generating comprehensive database descriptions, our reasoning steps generation module creates structured CoT reasoning steps that systematically guide models through the visualization creation process.
Following our identified five-stage CoT process, we decomposed the visualization reasoning into five key steps.
We leveraged GPT-4o-mini to \revise{complete the reasoning steps using the ground truth VQL and carefully crafted prompts, with full details provided in the supplemental material.
We selected GPT-4o-mini due to its strong reasoning capabilities and cost-effectiveness for large-scale data generation.}

\noindent\textbf{Determine chart type (S1)}.
This initial step analyzes the \sql{VISUALIZE} clause in the VQL output (\eg, bar, line, pie) to select the most appropriate chart type based on the user analysis intent and database schema. 
Models are required to justify why the chosen chart type effectively communicates the requested insights and fulfills the analysis intent.

\noindent\textbf{Retrieve relevant data (S2)}.
This step carefully identifies the necessary tables, columns, and conditions for the visualization.
By reasoning through the \sql{FROM}, \sql{SELECT}, and \sql{WHERE} clauses, only relevant data is extracted for further processing, aligning with the query.
In a similar way, models are required to justify each decision they make.

\noindent\textbf{Define data granularity (S3)}.
In this step, models are required to determine how to group and aggregate data, using \sql{GROUP BY} for categorical or numerical grouping and \sql{BIN BY} for time-based data.
It explains the grouping strategy to ensure the visualization accurately reflects trends or summaries in the data.

\noindent\textbf{Refine data for visualization (S4)}.
This step focuses on sorting and limiting operations, which correspond to \sql{ORDER BY}, \sql{SORT DIRECTION}, and \sql{LIMIT} clauses in VQL.
Models are required to explain why this refinement is needed and how they enhance readability or align with the analysis intent.

\noindent\textbf{Generate visualization (S5)}.
The final step synthesizes all previous reasoning results and generates the complete VQL.
This step ensures all components work harmoniously to produce a visualization that accurately addresses user needs while maintaining technical correctness.

After building these reasoning steps, \revise{we conducted tuning and validation experiments on the Llama3.1-8B-Instruct model} and found that sometimes the model will still generate illegal results, such as \sql{HISTOGRAM} in the \sql{VISUALIZE} field and unsupported functions like \sql{WEEKDAY(Date)} in the \sql{BIN BY} field.
Such issues can be addressed by providing explicit constraints in the input prompt.
These constraints include limiting visualization types to \sql{BAR}, \sql{PIE}, \sql{LINE}, and \sql{SCATTER} and restricting column selections to those in the database schema or valid derivations (\eg, \sql{COUNT}, \sql{AVG}, \sql{MAX}, \sql{MIN}).
Our ablation studies demonstrate that incorporating these constraints as text prompts in our training samples significantly improves model reasoning capabilities and VQL generation accuracy.

\subsection{\data}
Before automatically augmenting nvBench using the database description module and the reasoning steps generation module, 
\revise{we conducted a rule-based filtering and removed} 41 problematic samples, including duplicated queries (9), illegal VQLs (26), and empty VQLs (6).
Furthermore, we \revise{leveraged GPT-4o-mini to identify 1,351 samples exhibiting inconsistency between queries and VQLs}.
\revise{For example, consider the query ``What are the dates of the assessment notes, and count them by a bar chart.''}
The groundtruth VQL is \sql{Visualize BAR SELECT date\_of\_notes, COUNT(date\_of\_notes) FROM Assessment\_Notes BIN date\_of\_notes BY WEEKDAY}, which contains an erroneous \sql{BIN BY} clause that makes it inconsistent with the original query intent.
After removing these problematic samples, we conducted a comprehensive quality assurance process \revise{by randomly sampling 15\% of the augmented data and manually evaluating the appropriateness of the generated reasoning steps.}
Detailed statistics of the \data are presented in \revise{supplemental material}.

\section{\sys}

In addition to our dataset and \revise{pipeline}, we also developed \sys, a visual analysis tool that exposes detailed reasoning processes while enabling rich interactive exploration.
This tool bridges the gap between NL queries and visualization outputs by making the underlying CoT reasoning accessible and modifiable.

\subsection{Visualization Design}
Fig.~\ref{fig:interface} presents the interface of \sys.
The core component is the CoT view (Fig.~\ref{fig:interface}A), which provides a structured overview of the model's reasoning process from natural language queries to final visualization results.
The detailed reasoning steps are organized as a hierarchical tree following our five-stage CoT process.
The root node corresponds to S5, which synthesizes the previous reasoning steps and produces the final VQL result.
Placing this stage as the root allows users to quickly grasp the output before exploring the underlying reasoning process.
Four second-level nodes correspond to the four stages (S1-S4) in our CoT Process, each displaying the key decisions made by the model at that stage.
Their child nodes reveal more detailed reasoning steps, such as \sql{GROUP\_BY}, and \sql{BIN\_BY} fields inferences in S3.
To facilitate understanding without overwhelming the user with details, a concise summary of the reasoning is included in each node.
To optimize space usage while maintaining context, we implemented a space-tree layout with dynamic expansion controls.
This adheres to the ``details-on-demand'' principle, allowing users to selectively expand nodes of interest while collapsing others\revise{.}
This hierarchical visualization aligns with the top-down analysis workflow users naturally adopt, enabling them to efficiently navigate from high-level visualization to specific implementation details while maintaining contextual awareness throughout the exploration process.

In addition to the CoT view, we provided three supporting views to complement the CoT view during the analysis.
The information view (Fig.~\ref{fig:interface}B) provides comprehensive reasoning text for the selected step, which allows users to examine the detailed thought process.
The chart view (Fig.~\ref{fig:interface}C) renders the final visualization based on the model-generated VQL.
Users can also export the Vega-Lite specification or the SVG file once they confirm their satisfaction with the results.
For detailed data exploration, the table view (Fig.~\ref{fig:interface}D) provides a structured presentation of the underlying dataset, enabling users to verify how raw data translates into visual elements throughout the reasoning steps.

\subsection{Interactions}

\noindent\textbf{Coordinated exploration}.
We implemented coordinated view updates to maintain analysis context throughout the exploration process.
When users select a node within the CoT view, the information view instantly displays the comprehensive reasoning text associated with that step, and the table view shows the retrieved or transformed data after the selected steps.
These coordinated interactions create a cohesive analysis environment that maintains contextual continuity while navigating the complex reasoning chain.

\noindent\textbf{Interactive refinement}.
Beyond passive exploration, our interface allows users to actively refine the reasoning process when they identify potential errors or suboptimal decisions.
We offer two complementary refinement mechanisms:
\begin{itemize}[nosep]
    \item Self correction: This feature prompts the model to automatically reconsider its decisions within the selected reasoning step.
     \revise{By leveraging hints about potential errors in the flagged step}, the model can improve results without direct user input.
    \item Manual correction: This allows users to provide specific preferences that steer the regeneration process toward desired results, combining human domain expertise with model capabilities.
\end{itemize}
\revise{The corresponding prompts are provided in the supplemental material.}

After making corrections, our tool will intelligently regenerate all subsequent steps to ensure logical consistency throughout the reasoning process.
To help users understand the effects of their modifications, we implement an intuitive comparison feature that clearly identifies differences between original and revised reasoning paths.
As shown in Fig.~\ref{fig:interface}A, when a self-correction is applied to the \sql{SORT DIRECTION} step (Fig.~\ref{fig:interface}A${}_4$), unchanged nodes appear visually dimmed (\eg, Fig.~\ref{fig:interface}A${}_5$), while modified nodes are highlighted with affected fields marked in red for immediate identification (Figs.~\ref{fig:interface}A${}_1$-A${}_4$).
The newly generated reasoning step is appended under the node for \sql{SORT DIRECTION} (Fig.~\ref{fig:interface}A${}_6$), allowing users to directly compare the before-and-after reasoning processes and select the more appropriate one.

By making the CoT reasoning both transparent and interactive, our method transforms users from passive consumers of automated visualizations into active collaborators in the design process. 
This improves both the usability of \sys and the quality of resulting visualizations, ultimately leading to more effective data exploration and analysis.

\section{Evaluation}

\subsection{Quantitative Evaluation}
\subsubsection{Comparison with Baseline Methods}
\label{subsec:benchmark}
\noindent\textbf{Baseline methods}.
We chose \revise{seven} representative \task methods with different architectures for comparison.
\begin{itemize}[nosep]
    \item \textbf{Seq2Vis}~\cite{luo2021synthesizing}: This method treats the \task problem as a machine translation problem between natural language and visualization specifications.
    We included it \revise{because it is the pioneering work and establishes the foundational baseline.}
    \item \textbf{Transformer}: This is a milestone model and has been widely adopted for various NLP tasks.
    \revise{We included it to evaluate how a general-purpose NLP model performs on the \task task.}
    \item \textbf{ncNet}~\cite{luo2021ncnet}: 
    This is the state-of-the-art model for \task based on the Transformer architecture, which introduces several visualization-aware optimizations to better understand user intent and generate specification-compliant outputs.
    \revise{We included it to establish the current performance ceiling for specialized models.}
    \item \textbf{RGVisNet}~\cite{song2022rgvisnet}: This is an innovative hybrid retrieval-generation method that first retrieves the most relevant query candidates as prototypes from the VQL codebase and then revises them to produce the desired VQL.
    \revise{We included it to compare with different paradigms in \task and provide insights into whether our CoT methodology outperforms retrieval-augmented strategies.}
    \item \textbf{Llama3.1-8B-SFT}: \revise{This baseline employs the identical backbone architecture as our model, yet it adopts end-to-end data without CoT reasoning steps for supervised fine-tuning.
    We included it to directly validate the effectiveness of our CoT module.}
    \revise{\item \textbf{General Purpose LLMs}: We evaluate against seven state-of-the-art general purpose LLMs representing diverse architectures and capabilities: Llama3.1-8B (an open-source small-scale model), GPT-4o-mini (which also serves as the source for CoT reasoning steps generation in our method), GPT-o1 and GPT-o3 (OpenAI's latest reasoning models), Gemini-2.5-Pro (Google's flagship multimodal model), Claude-3.5-Sonnet (Anthropic's advanced reasoning model), and DeepSeek-R1 (an open-source reasoning model).
    We included them because they represent how general users perform \task using state-of-the-art LLMs.
    }
    \item\textbf{ChartGPT}~\cite{xie2024haichart}: This is the most comparable work \revise{that also breaks down the chart generation task into multiple steps: column selection, data filtering, data aggregation, chart type determination, and data visualization.
    We included it to highlight the effectiveness of our specific reasoning chain design.}
    
\end{itemize}
\noindent\textbf{Experiment settings}.
We adopted the split of the train/dev/test data set following \revise{Song \etal}\cite{song2022rgvisnet}, which achieves a strict separation of databases and ensures that no individual database appears across multiple sets.
This prevents potential data leakage and maintains the integrity of the evaluation. \revise{We selected Llama3.1-8B-Instruct as our base model due to its stable performance\cite{grattafiori2024llama}, open-source accessibility, and widespread adoption in recent research\cite{zhang2024sciinstruct,wang2024mamba}.}
The detailed hyperparameter settings are provided in the supplemental material. 

\noindent\textbf{Metrics}.
To comprehensively evaluate model performance, we compared the generated VQL with the groundtruth from multiple aspects.
Following RGVisNet\cite{song2022rgvisnet}, we measure accuracy in chart type (Chart Acc), x/y axes configuration (Axis Acc), and SQL syntax (SQL Acc), which are the three major components of VQL.
However, syntactically different SQL queries can produce identical results, such as the conditions \sql{WHERE YEAR IN (1999, 2000)} and \sql{WHERE YEAR = 1999 OR YEAR = 2000}.
Therefore, we introduce two additional metrics: Data Acc, which measures whether execution results match regardless of SQL syntax, and All Acc, which indicates when chart type, axes, and data all match correctly.
These execution-based metrics complement the syntax-based metrics for a more comprehensive assessment.

\begin{table}[t]
\setlength{\tabcolsep}{2pt}
    \centering
    \caption{\centering Performance Comparison.}
    \vspace{-3mm}
    \begin{tabular}{lccccc}
        \toprule
        Method & Chart Acc & Axis Acc & SQL Acc & Data Acc & All Acc  \\
        \midrule
        Seq2Vis & 93.18\% & 22.71\% & 0.80\% & 0.84\% & 0.62\% \\
        Transformer & 97.79\% & 62.24\% & 18.59\% & 18.73\% & 17.93\%  \\
        ncNet & \textbf{98.05\%} & 46.40\% & 42.59\% & 43.07\% & 42.28\%  \\
        RGVisNet & 97.21\% & 48.34\% & 53.61\% & 53.96\% & 51.70\%  \\
        \revise{Llama3.1-8B} &\revise{81.27\%} & \revise{68.75\%} & \revise{42.98\%} & \revise{51.62\%} & \revise{46.48\%}  \\
        Llama3.1-8B-SFT & 83.83\% & 77.05\% & 52.10\% & 62.52\% & 59.37\%  \\
        GPT-4o-mini & 91.31\% & 87.16\% & 52.35\% & 75.07\% & 70.37\%  \\
        \revise{GPT-o1} & \revise{91.39\%} & \revise{93.36\%} & \revise{59.76\%} & \revise{77.62\%} & \revise{71.63\%} \\
        \revise{GPT-o3} & \revise{92.70\%} & \revise{91.95\%} & \revise{59.39\%} & \revise{76.02\%} & \revise{71.14\%} \\
        \revise{Gemini-2.5-Pro} & \revise{92.03\%} & \revise{93.32\%} & \revise{46.57\%} & \revise{75.25\%} & \revise{69.81\%} \\
        \revise{Claude-3.7-Sonnet} & \revise{92.47\%} & \revise{93.98\%} & \revise{59.72\%} & \revise{77.64\%} & \revise{71.80\%} \\
        \revise{DeepSeek-R1} & \revise{92.25\%} & \revise{94.36\%} & \revise{51.75\%} & \revise{77.34\%} & \revise{72.60\%} \\
        ChartGPT  & 97.34\% & 94.85\% & 69.21\% & 73.84\% & 73.03\% \\
        \textbf{NL2VIS-CoT} & 97.52\% & \textbf{95.17\%} & \textbf{74.63\%} & \textbf{80.74\%} & \textbf{77.16\%} \\
        \bottomrule
    \end{tabular}
    \vspace{-3mm}
    \label{tab:benchmark}
\end{table}

\noindent\textbf{Results analysis}.
Table~\ref{tab:benchmark} summarizes the results of our proposed NL2VIS-CoT method and the baseline methods on the test set. Traditional models like Seq2Vis and Transformer exhibit high Chart Acc but struggle with SQL Acc and Data Acc.
This stems from their inability to effectively handle the hierarchical complex dependencies between natural language semantics and database schema parsing, which is critical for NL2VIS tasks.
In contrast, \revise{advanced specialized models like ncNet and RGVisNet address this through sophisticated design choices,} making them better interpret natural language queries and transform data accurately. 
Specifically, RGVisNet performs better than ncNet (51.70\% vs. 42.28\% All Acc), which can be attributed to its hybrid retrieval-generation framework that better utilizes existing VQL prototypes to reduce errors during visualization synthesis.
\revise{For LLM-based methods, Llama3.1-8B with few-shot prompting achieves 46.48\% in All Acc without any training, and Llama3.1-8B-SFT achieves higher performance (59.37\%) after supervised fine-tuning.
However, the performance is still constrained by model scale limitations.
Large-scale LLMs with few-shot prompting demonstrate better performance, and the results are consistent across different models: around 92\% in Chart Acc, 93\% in Axis Acc, 77\% in Data Acc, and 71\% in All Acc.}
\revise{However, they lag behind the traditional methods in terms of Chart Acc, revealing a critical limitation of few-shot learning: limited exemplars hinder effective generalization for different cases and may even generate invalid VQLs, such as \sql{VISUALIZE HISTOGRAM}.
Compared with previous LLM-based methods, ChartGPT and our method achieve higher Chart Acc and All Acc, highlighting the great potential of fine-tuning small-scale LLMs with CoT reasoning steps.}
Notably, our NL2VIS-CoT achieves a leading All Acc of 77.16\% \revise{while simultaneously performing best in Axis Acc, SQL Acc, and Data Acc.} 

\begin{figure*}[!t]
\includegraphics[width=\textwidth]{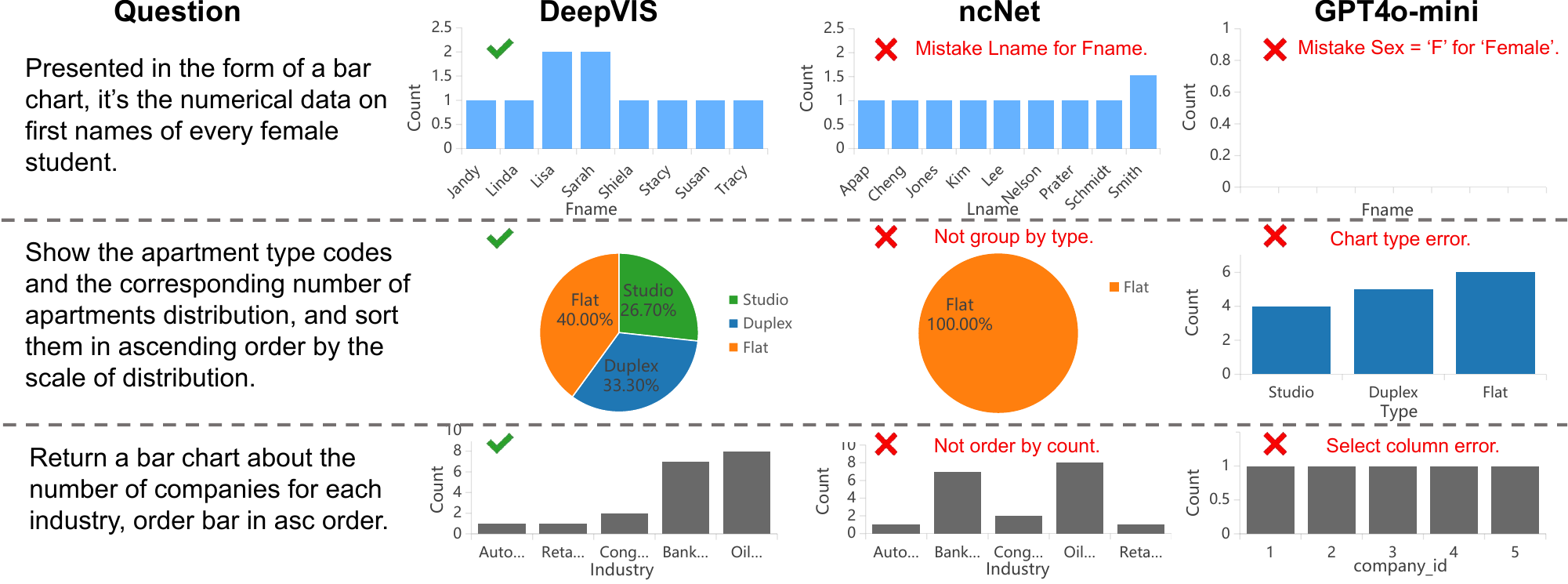}
\vspace{-3mm}
\caption{Comparative analysis of VQLs and generated charts between our method and baseline methods.}
\label{fig:baseline}
\end{figure*}

In addition to the numerical comparison, we also provide several examples to demonstrate why \model performs better than the baseline methods in Fig.~\ref{fig:baseline}.
Here we select ncNet and GPT-4o-mini as two representative methods for comparison, while the full results are provided in the supplemental material.

\subsubsection{Ablation Study}
\label{subsec:ablation}
In addition to the benchmark evaluation, we carry out ablation experiments to validate the effectiveness of our constructed \data and the significance of each component.
We consider four ablation methods:\looseness=-1

\begin{itemize}[nosep]
    \item \textbf{w/o value sampling}: We removed the column value sampling mechanism from the database description module to evaluate how this mechanism affect model performance.
    \item \textbf{w/o constraints}: We \revise{removed the explicit constraints in the input prompt} to understand how these constraints contribute to the model's behavior.
    \item \textbf{w/o CoT}: We removed the CoT reasoning process to analyze its role and importance in \task.
    \item \textbf{ChartGPT-pipeline}: ChartGPT also decomposes the generation of VQL into multiple steps but uses a different order from us.
    To evaluate the comparative effectiveness of reasoning step orders, we reconstructed our dataset using ChartGPT's proposed order and trained a model under identical experimental settings.
\end{itemize}

\noindent\textbf{Results analysis}.
Table~\ref{tab:ablation} shows the results of the ablation study.
When the value sampling mechanism was eliminated, Data Acc substantially dropped from 80.74\% to 72.11\%, with All Acc dropping from 77.16\% to 69.31\%.
This is mainly because the model fails to generate correct SQL queries to retrieve relevant data, particularly in the \sql{WHERE} clause.
For example, the model incorrectly generates \sql{WHERE rank="Asstant Professor"} instead of the correct one \sql{WHERE rank="AsstProf"}, resulting in a query that fails to retrieve any data.
The removal of constraints produced a different pattern of performance degradation.
While SQL Acc and Data Acc show relatively modest declines compared to removing value sampling, Chart Acc and Axis Acc exhibit more significant drops.
This emphasizes the importance of constraints in generating reasonable results.
In one example, the model incorrectly uses a function \sql{WEEKDAY()} instead of the standardized \sql{BIN BY} syntax, resulting in an illegal VQL.
When removing the detailed CoT reasoning steps, all the metrics significantly drop, indicating the critical importance of the reasoning process in our method.
Compared with the ChartGPT-pipeline, our method also achieves better performance across all metrics, empirically validating the effectiveness of our order.

\begin{table}[h]
\setlength{\tabcolsep}{1.5pt}
    \centering
    
    \caption{\centering Ablation Study Results.}
    \vspace{-3mm}
    \begin{tabular}{lccccc}
        \toprule
        Method & Chart Acc & Axis Acc & SQL Acc & Data Acc & All Acc \\
        \midrule
        w/o value sampling & 94.99\% & 88.95\% & 67.85\% & 72.11\% & 69.31\%  \\
        w/o constraints & 94.16\% & 87.03\% & 71.81\% & 78.23\% & 75.39\% \\
        w/o CoT & 70.30\% & 65.79\% & 48.30\% & 51.88\% & 49.31\% \\
        ChartGPT pipeline & 95.60\% & 93.08\% & 71.10\% & 77.32\% & 73.54\%\\
        \midrule
        \textbf{\model} & \textbf{97.52\%} & \textbf{95.17\%} & \textbf{74.63\%} & \textbf{80.74\%} & \textbf{77.16\%}  \\
        \bottomrule
        \label{tab:ablation}
    \end{tabular}
    \vspace{-3mm}
\end{table}

\revise{\subsubsection{Error Analysis}}
\label{subsec:Error}
\label{subsec:error}
\revise{\noindent 
To better understand the limitations of our method and identify areas for improvement, we conducted a comprehensive error analysis across our four-step reasoning.
Our analysis reveals that S1 (determine chart) had fewer errors (56), while S2 (retrieve data), S3 (define granularity), and S4 (refine data) exhibited much more errors (183, 202, and 253 samples, respectively).
Notably, these included 131 aggregation function errors in S2, 191 \sql{GROUP BY} errors in S3, and 220 \sql{ORDER BY} errors in S4.
For example, when processing the query ``Can you draw the trend of maximal score over the year? rank by the x-axis in descending,'' our method generates \sql{VISUALIZE LINE SELECT YEAR, MAX(SCORE) FROM WINE ORDER BY YEAR DESC}.
While the system correctly identified the aggregation function \sql{MAX(SCORE)} and the sorting requirement, it failed to include the essential \sql{GROUP BY YEAR} clause necessary for proper aggregation.
This demonstrates that while our model effectively handles common queries, there remains room for improvement in complex queries involving data aggregation and transformation logic.
}

\revise{
Next, we focused our analysis on S1 due to its foundational role in the reasoning pipeline and its cascading impact on subsequent steps.
Our findings reveal significant variations in error rates across different chart types: \sql{BAR} (1.18\%), \sql{PIE} (2.97\%), \sql{SCATTER} (7.63\%), and \sql{LINE} (9.83\%).
Notably, despite pie charts representing only 7.84\% of the dataset, they maintain a relatively low error rate, suggesting that chart frequency does not directly correlate with prediction accuracy.
To better understand the high error rates in line charts and scatter plots, we conducted a deeper investigation into these errors and found that over 95\% of them occur in multi-solution scenarios, \ie, multiple chart types can effectively address the same question. 
For example, when asked ``How many documents correspond with each project id,'' the ground truth specifies a scatter plot, yet our model's bar chart response equally solves the problem.
This pattern exposes a limitation in the current dataset evaluation framework, which fails to account for multiple valid solutions corresponding to a single natural language query.
}
\vspace{2mm}

\subsection{Use Cases}
We present two use cases to demonstrate how our interactive interface enhances user understanding of the reasoning process and enables targeted adjustments.

\begin{figure*}[!t]
\includegraphics[width=\textwidth]{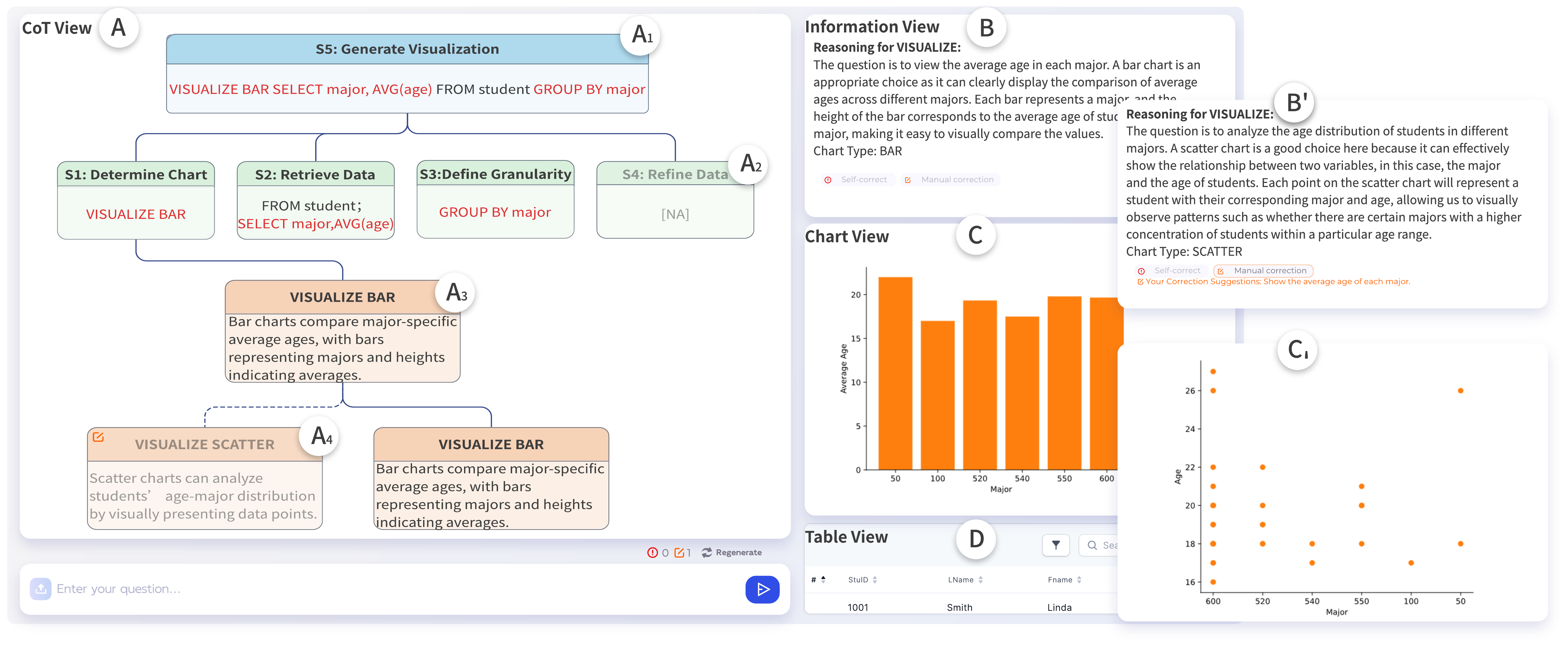}
\vspace{-4mm}
\caption{After using manual correction to specify the preference of the chart type, \sys correctly updates the subsequent reasoning results and generates a proper aggregated visualization.}
\label{fig:case2}
\end{figure*}

\subsubsection{Case 1: Using Self Correction to Refine Results.}
This use case illustrates how users can explore reasoning steps and guide the model to reconsider critical decisions, leading to enhanced visualization outcomes.

Alice analyzes the allergy database, which contains a table \sql{student} with columns such as \sql{stuid}, \sql{name}, \sql{city\_code}, and \sql{age}.
She aims to examine student distribution across cities and identify the city with the most students.
Therefore, she inputs ``Please display a bar chart showing all cities and their corresponding number of students to identify the city with the highest student count.''
Fig.~\ref{fig:interface}C' displays the generated chart, which successfully uses a bar chart to visualize student numbers by city.
Alice verifies the reasoning process and confirms it is sound. 
For example, Fig.~\ref{fig:interface}B' shows that the model appropriately justifies the chart type selection: ``A bar chart (\sql{BAR}) is a perfect fit as it allows for a clear comparison of the student counts across different cities.''
However, while the chart enables identification of the city with the most students, Alice prefers to sort the data to make the pattern more evident. 
The model initially overlooked sorting because it reasoned: ``Since the question doesn't demand result sorting, we omit the \sql{ORDER BY} clause'' (Fig.~\ref{fig:interface}A${}_6$).
To address this issue, Alice activates the ``Self correct'' feature to prompt the model to reconsider this decision.
Upon reconsideration, the model now acknowledges that ``Sorting the data in descending order based on student count will place the city with the maximum number of students at the top, making it straightforward to visually determine which city has the highest student count'' (Fig.~\ref{fig:interface}B), and
the node for \revise{reasoning} \sql{SORT DIRECTION} also updates accordingly (Fig.~\ref{fig:interface}A${}_4$).
Fig.~\ref{fig:interface}\revise{C} shows the improved visualization with sorted bars, which Alice finds satisfactory.

\begin{figure*}[t]
    \includegraphics[width=\textwidth]{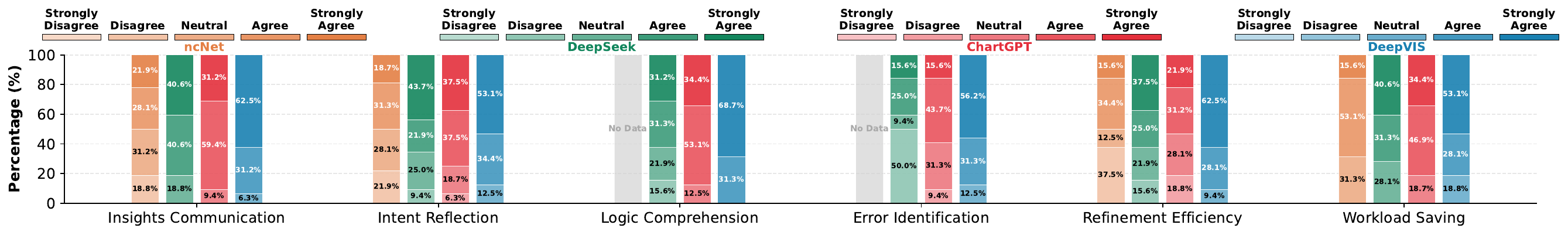}
    \vspace{-4mm}
    \caption{User study results: \sys has consistently received relatively higher ratings across all six dimensions.}
    \label{fig:scores}
\end{figure*}

\subsubsection{Case 2: Use Manual Correction to Refine Results.}
This use case illustrates how users can explore reasoning steps and make target refinements to their query.

Bob is tasked with analyzing a university database to explore the age distribution of students across different majors.
The database includes a \sql{student} table with key columns such as \sql{stuid}, \sql{age}, \sql{sex}, \sql{major}, \sql{advisor}, and \sql{city\_code}, alongside other tables like \sql{faculty} and \sql{department}. 
To begin his analysis, he inputs the query: ``Analyze the age distribution of students in different majors'' into \sys.
\sys responds by generating a scatter plot with individual data points for each student's major and age (Fig.~\ref{fig:case2}C').
He finds this scatter plot difficult to interpret and compare the age distribution across different majors.
Therefore, he decides to examine the reasoning process to see why \sys select scatter plot.
After clicking the node for reasoning the chart type, he finds that \sys thinks a scatter plot effectively displays the relationship between two variables—major and age, revealing patterns like age concentrations within specific majors (Fig.~\ref{fig:case2}A${}_4$).
He then realizes that his query may not be so accurate and may mislead the model.
Therefore, he refines his requirement to ``Show the average age of each major'' and clicks ``Manual correction.''
After applying this correction, \sys changes the chart type from \sql{SCATTER} to \sql{BAR}, selection from`\sql{age} to \sql{AVG(age)}, and adds \sql{GROUP\_BY major} to aggregate the data by major.
The resulting bar chart displays each major as a distinct bar, with the height of each bar representing the average age of students in that major (Fig.~\ref{fig:case2}C).
This new visualization proves far clearer than the initial scatter plot, enabling Bob to easily compare the central tendency of ages across different majors.
Reflecting on the process, Bob realizes that his original query ``Analyze the age distribution'' lacks specificity.
However, the reasoning steps help him realize it and successfully guide the tool to produce a more suitable output.
He also appreciates that \sys automatically adjusted related fields to align with his corrected request, sparing him the effort of modifying each field manually.

\subsection{User Study}
We conducted a user study to evaluate whether presenting reasoning steps in \sys facilitates users in understanding model behavior and generating better visualization.

\noindent\textbf{Participants}.
We recruited \revise{32} participants (\revise{P1-P32, 16} males and \revise{16} females) for the experiment, \revise{20 are from the local university and 12 are from the workforce.}
including 20 from the local university \revise{and 12 industry professionals, 20–51 years old (M=25.78, SD=7.39)}.
They are from diverse majors, including Computer Science (\revise{8}), Software Engineering (\revise{5}), Finance (\revise{5}), Mathematical Statistics (3), Digital Media Design (2), Journalism (2), \revise{Civil Engineering (3), Biomedical Engineering (2)} and Geographic Information Science (2). 
All participants have experience using data visualization tools, such as Excel, Matplotlib, ECharts, and Vega-Lite. 
In addition, \revise{25} of them have tried using LLMs to assist in data visualization.

\noindent\noindent\textbf{Study procedure}.
We began the user study by introducing \task tasks using representative nvBench examples.
Then, participants were introduced to \revise{four interfaces:} ncNet (using the developers' Jupyter notebook), \revise{DeepSeek and ChartGPT (using chatbot interfaces integrating the backend models)}, and our \sys.
Following familiarization, participants completed 10 randomly sampled nvBench examples across all interfaces, with the order counterbalanced to mitigate learning and fatigue effects.
Upon task completion, participants responded to a five-point Likert-scale questionnaire to evaluate the effectiveness and usability of the interfaces.
Finally, we conducted a brief interview with participants to collect detailed feedback \revise{and analyzed the interview data using thematic analysis~\cite{braun2012thematic}, where the first author did the initial coding and revised it with an external second coder.}\looseness=-1

\noindent\textbf{Result analysis}.
Fig.~\ref{fig:scores} shows the rating across six perspectives: insights communication, intent reflection, logic comprehension, error identification, refinement efficiency, and workload saving.
The detailed questions are provided in the supplemental material.
Overall, \sys has consistently received a relatively high proportion of ``Strongly Agree'' and ``Agree'' ratings across all six dimensions, which provides compelling evidence of its effectiveness in enhancing user understanding and facilitating interactive refinement.
P7 commented positively regarding the reasoning steps generated by the \sys: ``\textit{These reasoning steps sound reasonable and justify the choice made by models, making the visualization decisions transparent and understandable.}''
Other participants also agreed that the detailed reasoning steps helped them ``\textit{identify the errors in reasoning more easily}'', (P2) ``\textit{allow for quicker iteration}'' (P8), and ``\textit{provide useful hints to refine the input query}'' (P11).
Such feedback highlights the benefits of disclosing the reasoning steps in the analysis.
With respect to overall workload reduction, the participants consistently acknowledged the effectiveness of \sys.
Many expressed sentiments similar to those of P4, who remarked: ``\textit{It alleviates the tasks of creating visualization by automatically handling data transformation and visual encoding}.''
P1 further added: ``\textit{I would like to use this tool in the future for my data analysis projects},'' demonstrating the practical usability of \sys.

In comparative evaluations against \revise{other baseline methods}, \sys exhibited better performance across all perspectives.
A particularly notable distinction emerged in refinement efficiency, \revise{where our method received only 9\% negative rating,  substantially lower than ncNet (50\%), DeepSeek (38\%), and ChartGPT (47\%).}
This performance advantage stems from our transparent interface design, which exposes intermediate reasoning steps and enables targeted interactive refinement.
In contrast, other interfaces require users to restart the entire process when modifications are needed. 
For example, P2 commented on ncNet: ``\textit{While this package is very easy to use, it only generates the final visualization, and I sometimes need to try different queries to obtain a satisfactory result, which can be time-consuming and frustrating.}''
Regarding the workload saving, P17 pointed out that she needs to ``\textit{provide more detailed instructions to steer ncNet {and DeepSeek} compared to \sys.}''
This highlights the advantage of \sys in reducing the cognitive burden on users while producing high-quality visualizations that accurately reflect their analysis intent.

\section{Discussion and Future work}
Based on the interview with our experts and the participants in the user study, we discuss several promising directions for future work.

\noindent\textbf{Integrating visualization feedback}. While our CoT process has shown promising results, it currently lacks direct integration of the final visualization or its underlying data into the reasoning process.
This limitation can result in suboptimal visualizations, especially when multiple alternatives could meet user demands.
For example, while both bar charts and scatter plots can reveal relationships between variables, the optimal choice depends on data characteristics.
Scatter plots work better with fewer data points, while bar charts are preferable when presenting aggregated values..
To address this limitation, we suggest integrating feedback from the final visualization into the creation process by adapting our CoT pipeline.
We propose two key strategies to achieve this: First, we could explore leveraging multi-modal large language models to analyze generated visualizations and provide actionable insights, such as detecting visual clutter or recommending alternative chart types.
Second, tools like VizLinter~\cite{chen2021vizlinter} can automatically pinpoint common visualization flaws.
These flaws can be described in natural language and integrated into the reasoning process, allowing models to suggest fixes or apply corrections automatically.

\noindent\textbf{Enhancing fine-grained control}.
Currently, we translate the natural language queries into VQLs, which are then converted into Vega-Lite specifications for rendering.
This design choice abstracts away rendering-specific details, enabling cross-tool compatibility with visualization frameworks such as Vega-Lite, ECharts, and matplotlib.
However, this also limits fine-grained control on the chart, such as setting color scales, adjusting layout tuning, and changing mark size.
This issue can be addressed by further advancing our CoT framework.
On the one hand, we can analyze existing visualization tools, summarize common specifications, and incorporate additional fields, such as \sql{COLOR\_BY} for color encoding and \sql{Mark\_Size} controlling mark dimensions.
This enhancement would allow users to specify detailed aesthetic preferences while preserving the benefits of cross-tool compatibility.
On the other hand, we can explore datasets that map natural language queries directly to Vega-Lite specifications, \revise{which supports more diverse chart types and allows more detailed configurations.}
The dataset can be constructed in a similar way, \ie, prompting the LLMs to generate detailed, step-by-step reasoning traces that connect analysis intent to ground truth specifications.
The generated reasoning steps can guide models in learning how to achieve analysis intent using Vega-Lite.

\noindent\textbf{Extending the CoT framework to broader tasks}.
By incorporating CoT reasoning into the \task pipeline, we have demonstrated how structured reasoning steps can improve both model performance and transparency in \task.
\revise{Our method is highly flexible and can be readily adapted to more complex \task datasets.}
\revise{For example, a preprint dataset nvBench2.0~\cite{nvbench2} extends nvBench by providing multiple valid VQLs for identical queries.
On the one hand, our method can directly generate intermediate reasoning steps for this dataset without modification due to their similar data format.
On the other hand, users can make targeted modifications to the prompts in our reasoning steps generation module, which instructs models to produce multiple valid outputs at each reasoning step and generate diverse VQLs.
Furthermore, we would like to highlight that by augmenting existing datasets with reasoning steps, models are able to solve a complex task beyond \task.
}
For example, in data storytelling, it can help explain the reasoning behind the visual choices so that the model learns how to better connect the stories to visuals.
In visualization debugging tasks, the reasoning process provides rich contextual insights into why certain visualizations may fail, allowing models to refine design choices more effectively.
Future work can explore how this framework facilitates various visualization tasks and achieves better human-AI collaboration.
\section{Conclusion}
To tackle the challenges that existing \task methods suffer from a lack of transparency and are challenging to refine due to their black-box designs, we propose the integration of the CoT process into the \task pipeline.
Our work delivers three key contributions: 1) designing a comprehensive CoT reasoning process for \task, 2) introducing the \data dataset with detailed reasoning steps to achieve state-of-the-art performance, and 3) developing an interactive visual interface that lets users inspect and tweak the reasoning behind visualizations.
Quantitative benchmark evaluation and qualitative case studies demonstrate that our method outperforms traditional methods, with users appreciating the inherent transparency that fosters trust and expertise.
Furthermore, our method suggests the broader potential of CoT reasoning to enhance model performance and foster effective human-AI collaboration across diverse visualization tasks.

\section*{Supplemental Materials}
\label{sec:supplemental_materials}
All the supplemental materials are available at the website \href{https://github.com/Bvivib-shuai/DeepVIS}{https://github.com/Bvivib-shuai/DeepVIS}, including: 1) a PDF file including the prompts for the reasoning steps generation module, exemplar natural language queries, VQLs, and visualizations generated by baseline methods and our methods, and the questionnaire for the user study, 2) a video demonstrating the interface and the two use cases, 3) the \data dataset, 4) the code for training data and deploying \sys, \revise{5) the prompt of interactive refinement, and 6) implementation details.}

%% if specified like this the section will be omitted in review mode
\acknowledgments{%
This work is partly supported by the NSF of China (62402409), the Guangdong Basic and Applied Basic Research Foundation (2023A1515110545), the Guangzhou Basic and Applied Basic Research Foundation
(2025A04J3935), and the Guangzhou-HKUST(GZ) Joint Funding Program (2025A03J3714).%
}

\bibliographystyle{abbrv-doi-hyperref}

\bibliography{template}

\appendix % You can use the `hideappendix` class option to skip everything after \appendix

\end{document}